\documentclass[10pt,twocolumn,letterpaper]{article}
\usepackage{cvpr}
\usepackage{times}
\usepackage{epsfig}
\usepackage{graphicx}
\usepackage{caption}
\usepackage{amsmath}
\usepackage{amssymb}
\usepackage{color}
\usepackage{authblk}

\usepackage[breaklinks=true,bookmarks=false]{hyperref}

\cvprfinalcopy % *** Uncomment this line for the final submission

\def\cvprPaperID{3151} % *** Enter the CVPR Paper ID here

\begin{document}

%%%%%%%%% TITLE

% \author{Dae Young Park\textsuperscript{1}{*}\\ 
%Artificial Intelligence Research Institute, Korea\\
% {\tt\small likebullet86@gmail.com}
% \and
% Kwang Hee Lee\textsuperscript{2}{*}\\
%Boeing Korea Engineering and Technology Center\\
% {\tt\small lkwanghee@gmail.com}\\
% }

% \author{Dae Young Park\textsuperscript{*}\\ 
% AIRI\textsuperscript{1}, Korea\\
% {\tt\small likebullet86@gmail.com}
% \and
% Kwang Hee Lee\textsuperscript{*}\\
% BKETC\textsuperscript{2}\\
% {\tt\small lkwanghee@gmail.com}
% }

% \author[]{Dae Young Park}
% \author[]{Kwang Hee Lee}
% \affil[1]{Artificial Intelligence Research Institute, Korea}
% \affil[2]{Boeing Korea Engineering and Technology Center}
    
\title{Arbitrary Style Transfer with Style-Attentional Networks }

\author[1,*]{Dae Young Park}
\author[1,*]{Kwang Hee Lee}
\affil[1]{Artificial Intelligence Research Institute, Korea}
%\affil[2]{Boeing Korea Engineering and Technology Center}
\affil[ ]{\tt\small \textsuperscript{1}likebullet86@gmail.com, \textsuperscript{2}lkwanghee@gmail.com}

\renewcommand\Authands{ and }
  
\twocolumn[{%
\maketitle
\renewcommand\twocolumn[1][]{#1}%
\begin{center}
    \centering
    \begin{center}
    \includegraphics[width=0.85\linewidth,height=0.51\linewidth]{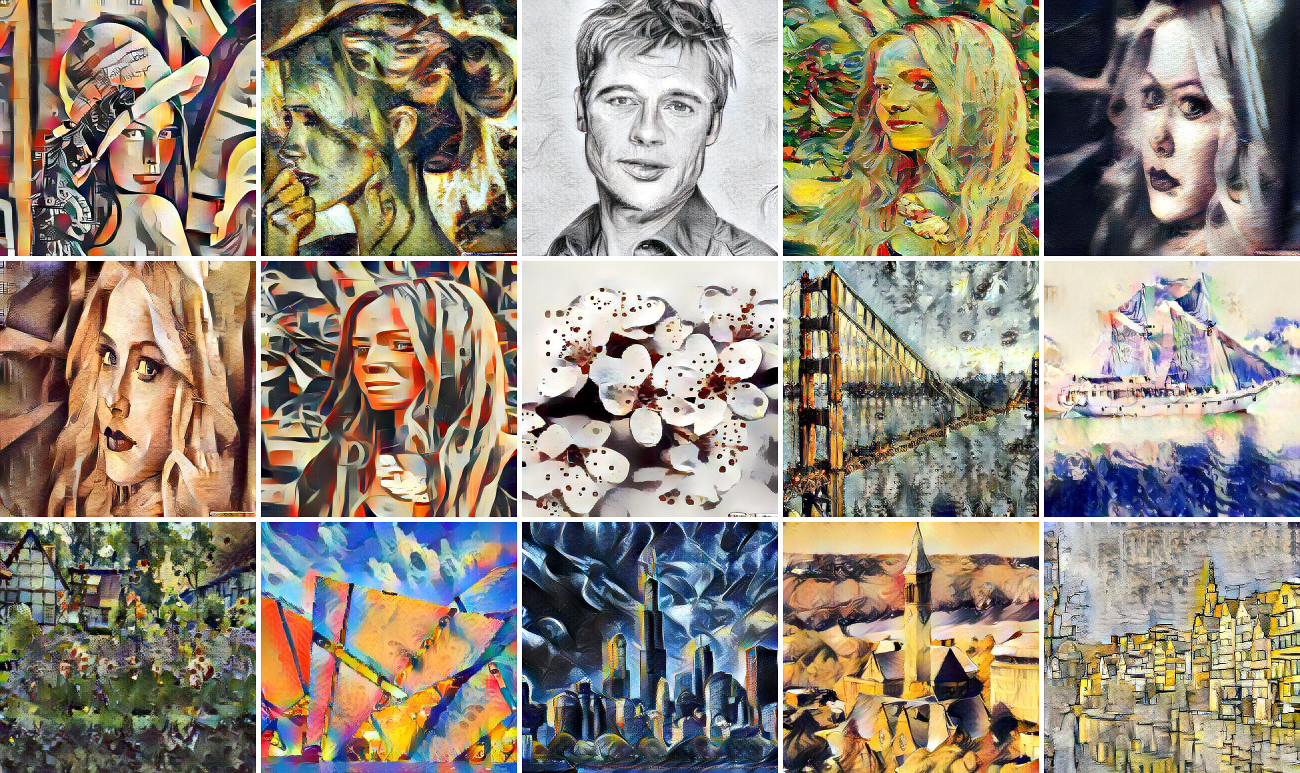}
    \end{center}
    \captionof{figure}{Result of the proposed SANet. We can transfer various styles to content images with high-quality.}
    \label{fig:head}
\end{center}%
}]

\newcommand\blfootnote[1]{%
  \begingroup
  \renewcommand\thefootnote{}\footnote{#1}%
  \addtocounter{footnote}{-1}%
  \endgroup
}

\blfootnote{ \textsuperscript{*} indicates equal contribution}
%%%%%%%%% ABSTRACT
\begin{abstract}
  Arbitrary style transfer aims to synthesize a content image with the style of an image to create a third image that has never been seen before. Recent arbitrary style transfer algorithms find it challenging to balance the content structure and the style patterns. Moreover, simultaneously maintaining the global and local style patterns is difficult due to the patch-based mechanism. In this paper, we introduce a novel style-attentional network (SANet) that efficiently and flexibly integrates the local style patterns according to the semantic spatial distribution of the content image. A new identity loss function and multi-level feature embeddings enable our SANet and decoder to preserve the content structure as much as possible while enriching the style patterns. Experimental results demonstrate that our algorithm synthesizes stylized images in real-time that are higher in quality than those produced by the state-of-the-art algorithms.
\end{abstract}

% \textsuperscript{*}

%%%%%%%%% BODY TEXT

\section{Introduction}

Artistic style transfer is a technique used to create art by synthesizing global and local style patterns from a given style image evenly over a content image while maintaining its original structure. Recently, the seminal work of Gatys et al. \cite{gatys2016image} showed that the correlation between features extracted from a pre-trained deep neural network can capture the style patterns well. The method by Gatys et al. \cite{gatys2016image} is flexible enough to combine the content and style of arbitrary images, but is prohibitively slow due to the iterative optimization process.

Significant efforts have been made to reduce the computational cost of this process. Several approaches \cite{chen2017stylebank, johnson2016perceptual, li2017diversified, ulyanov2016texture, dumoulin2017learned, li2017demystifying, shen2017meta, wang2017zm, zhang2017multi} have been developed based on feedforward networks. Feedforward methods can synthesize stylized images efficiently, but are limited to a fixed number of styles or provide insufficient visual quality. 

For arbitrary style transfer, a few methods \cite{li2017universal, huang2017arbitrary, sheng2018avatar} holistically adjust the content features to match the second-order statistics of the style features. AdaIN \cite{huang2017arbitrary} simply adjusts the mean and variance of the content image to match those of the style image. Although AdaIN effectively combines the structure of the content image and the style pattern by transferring feature statistics, its output suffers in quality due to the over-simplified nature of this method. WCT \cite{li2017universal} transforms the content features into the style feature space through a whitening and coloring process with the covariance instead of the variance. By embedding these stylized features within a pre-trained encoder--decoder module, the style-free decoder synthesizes the stylized image. However, if the feature has a large number of dimensions, WCT will accordingly require computationally expensive operations. Avatar-Net \cite{sheng2018avatar} is a patch-based style decorator module that maps the content features with the characteristics of the style patterns while maintaining the content structure. Avatar-Net considers not only the holistic style distribution, but also the local style patterns. However, despite valuable efforts, these methods still do not reflect the detailed texture of the style image, distort content structures, or fail to balance the local and global style patterns.

In this work, we propose a novel arbitrary style transfer algorithm that synthesizes high-quality stylized images in real time while preserving the content structure. This is achieved by a new style-attentional network (SANet) and a novel identity loss function. For arbitrary style transfer, our feedforward network, composed of SANets and decoders, learns the semantic correlations between the content features and the style features by spatially rearranging the style features according to the content features.

Our proposed SANet is closely related to the style feature decorator of Avatar-Net \cite{sheng2018avatar}. 
% However, the biggest difference between both approaches is that the SANet can flexibly decorate the style features by learning through the conventional style reconstruction loss and identity loss. In addition, our identity loss function helps the SANet maintain as much of their original content structure as possible through the keeping diversity of global and local style patterns.
There are, however, two main differences: The proposed model uses 1) a learned similarity kernel instead of a fixed one and 2) soft attention instead of hard attention. 
In other words, we changed the self-attention mechanism to a learnable soft-attention-based network for the purpose of style decoration. Our SANet uses the learnable similarity kernel to represent the content feature map as a weighted sum of style features that are similar to each of its positions. Using the identity loss during the training, the same image pair are input and our model is trained to restore the same result. At inference time, one of the input images is replaced with the style image, and the content image is restored as much as possible based on the style features. Identity loss, unlike the content--style trade-off, helps to maintain the content structure without losing the richness of the style because it helps restore the contents based on style features.
The main contributions of our work are as follows:

$\bullet$ We propose a new SANet to flexibly match the semantically nearest style features onto the content features.

$\bullet$ We present a learning approach for a feedforward network composed of SANets and decoders that is optimized using a conventional style reconstruction loss and a new identity loss.

$\bullet$ Our experiments show that our method is highly efficient (about 18--24 frames per second (fps) at 512 pixels) at synthesizing high-quality stylized images while balancing the global and local style patterns and preserving content structure.

\begin{figure*}
\begin{center}
\includegraphics[width=1\linewidth, height=0.3\linewidth]{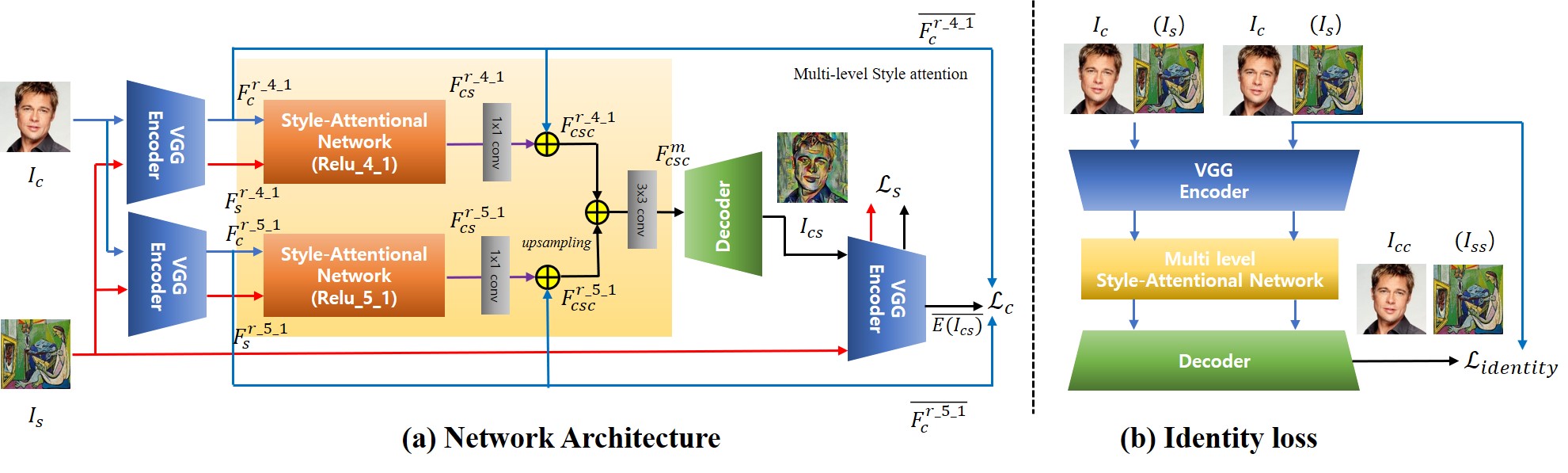}
\end{center}
  \caption{Overview of training flow. (a) Fixed VGG encoder encoding content and style images. Two SANets map features from \texttt{Relu\_4\_1} and \texttt{Relu\_5\_1} features respectively. The decoder transforms the combined SANet output features to $I_{cs}$ (Eq. \ref{eq:ics}). The fixed VGG encoder is used to compute $\mathcal{L}_{c}$ (Eq. \ref{eq:loss_content}) and $\mathcal{L}_{s}$ (Eq. \ref{eq:loss_style}). (b) The identity loss $\mathcal{L}_{identity}$ (Eq. \ref{eq:loss_identity}) quantifies the difference between $I_{c}$ and $I_{cc}$ or between $I_{s}$ and $I_{ss}$, where $I_{c}$ ($I_{s}$) is the original content (style) image and  $I_{cc}$ ($I_{ss}$) is the output image synthesized from the image pair (content or style).}
\label{fig:overview}
\end{figure*}

\section{Related Work}
{\bf Arbitrary Style Transfer.} The ultimate goal of arbitrary style transfer is to simultaneously achieve and preserve generalization, quality, and efficiency. Despite recent advances, existing methods \cite{gatys2016image, gatys2015texture, chen2017stylebank, johnson2016perceptual, li2017diversified, ulyanov2016texture, dumoulin2017learned, gatys2017controlling, li2016combining, li2016precomputed, ulyanovinstance, ulyanov2017improved, wang2017multimodal, risser2017stable} present a trade-off among generalization, quality, and efficiency.
Recently, several methods \cite{li2017universal, sheng2018avatar, chen2016fast, huang2017arbitrary} have been proposed to achieve arbitrary style transfer. The AdaIN algorithm simply adjusts the mean and variance of the content image to match those of the style image by transferring global feature statistics. WCT performs a pair of feature transforms, whitening and coloring, for feature embedding within a pre-trained encoder-decoder module. Avatar-Net introduced the patch-based feature decorator, which transfers the content features to the semantically nearest style features while simultaneously minimizing  the difference between  their holistic feature distributions.
In many cases, we observe that the results of WCT and Avatar-Net fail to sufficiently represent the detailed texture or maintain the content structure. We speculate that WCT and Avatar-Net could fail to synthesize the detailed texture style due to their pre-trained general encoder--decoder networks, which are learned from general images such as MS-COCO datasets \cite{lin2014microsoft} with large differences in style characteristics. As a result, these methods consider mapping the style feature onto the content feature in the feature space, but there is no way to control the global statistics or content structure of the style. Although Avatar-Net can obtain the local style patterns through a patch-based style decorator, the scale of style patterns in the style images depends on the patch size. Therefore, the global and local style patterns cannot both be taken into consideration. In contrast, AdaIN transforms texture and color distribution well, but does not represent local style patterns well. In this method, there exists another trade-off between content and style due to a combination of scale-adapted content and style loss. In this paper, we try to solve these problems using the SANets and the proposed identity loss. In this way, the proposed style transfer network can represent global and local style patterns and maintain the content structure without losing the richness of the style.  

{\bf Self-Attention Mechanism.} Our style-attentional module is related to the recent self-attention methods \cite{vaswani2017attention, zhang2018self} for image generation and machine translation. These models calculate the response at a position in a sequence or an image by attending to all positions and taking their weighted average in an embedding space. The proposed SANet learns the mapping between the content features and the style features by slightly modifying the self-attention mechanism.
%-------------------------------------------------------------------------
%-------------------------------------------------------------------------
\section{Method}
The style transfer network proposed in this paper is composed of an encoder--decoder module and a style-attentional module, as shown in Fig.~\ref{fig:overview}. The proposed feedforward network effectively generates high-quality stylized images that appropriately reflect global and local style patterns. Our new identity loss function helps to maintain the detailed structure of the content while reflecting the style sufficiently.

%-------------------------------------------------------------------------
\subsection{Network Architecture}

Our style transfer network takes a content image $I_c$ and an arbitrary style image $I_s$ as inputs, and synthesizes a stylized image $I_{cs}$ using the semantic structures from the former and characteristics from the latter. In this work, the pretrained VGG-19 network \cite{simonyan2014very} is employed as encoder and a symmetric decoder and two SANets are jointly trained for arbitrary style transfer. Our decoder follows the settings of \cite{huang2017arbitrary}.

To combine global style patterns and local style patterns adequately, we integrate two SANets by taking the VGG feature maps encoded from different layers (\texttt{Relu\_4\_1} and \texttt{Relu\_5\_1}) as inputs and combining both output feature maps. From a content image $I_c$ and style image $I_s$ pair, we first extract their respective VGG feature maps $F_c=E(I_c)$ and $F_s=E(I_s)$ at a certain layer (e.g., \texttt{Relu\_4\_1}) of the encoder.

After encoding the content and style images, we feed both feature maps to a SANet module that maps the correspondences between the content feature map $F_c$ and the style feature map $F_s$, producing following the output feature map:
\begin{equation}
\begin{split} 
{F}_{cs}&={SANet}\left({F}_{c},{F}_{s}\right)
\end{split} 
\label{eq:fcs}
\end{equation}

After applying 1 $\times$ 1 convolution to $F_{cs}$ and summing the two matrices element-wise as follows, we obtain $F_{csc}$:
\begin{equation}
\begin{split} 
{F}_{csc}&={F}_{c}+{W}_{cs}{F}_{cs},
\end{split} 
\label{eq:fcsc}
\end{equation}
where ``+" denotes element-wise summation.

We combine two the output feature maps from the two SANets as
\begin{equation}
\begin{split} 
{F}_{csc}^m&={conv}_{3\times3}({F}_{csc}^{\texttt{r\_4\_1}}+{upsampling}({F}_{csc}^{\texttt{r\_5\_1}})),
\end{split} 
\label{eq:multi_feature}
\end{equation}
where $F_{csc}^{\texttt{r\_4\_1}}$ and $F_{csc}^{\texttt{r\_5\_1}}$ are the output feature maps obtained from the two SANets, ${conv}_{3\times3}$ denotes the 3${\times}$3 convolution used to combine the two feature maps, and $F_{csc}^{\texttt{r\_5\_1}}$ is added to $F_{csc}^{\texttt{r\_4\_1}}$ after upsampling.

Then, the stylized output image $I_{cs}$ is synthesized by feeding $F_{csc}^{m}$ into the decoder as follows:
\begin{equation}
\begin{split} 
{I}_{cs}&=D\left({F}_{csc}^m\right).
\end{split} 
\label{eq:ics}
\end{equation}

%-------------------------------------------------------------------------
\subsection{SANet for Style Feature Embedding}
 Figure~\ref{fig:sanet} shows style feature embedding using the SANet module. Content feature maps $F_c$ and style feature maps $F_s$ from the encoder are normalized and then transformed into two feature spaces $f$ and $g$ to calculate the attention between $\overline{F_c^i}$ and $\overline{F_s^j}$ as follows:

\begin{figure}
\begin{center}
\includegraphics[width=0.9\linewidth, height=0.5\linewidth]{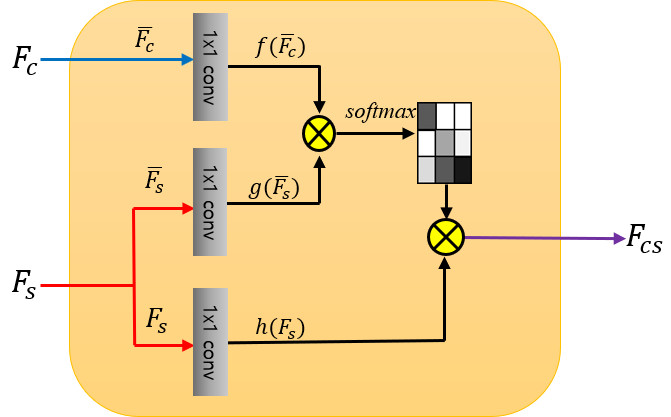}
\end{center}
  \caption{SANet.}
\label{fig:sanet}
\end{figure}

\begin{equation}
\begin{split} 
{F}_{cs}^i&=\frac{1}{C({F})}\sum_{\forall{j}}\exp({f}(\overline{{F}_{c}^i})^T{g}(\overline{{F}_{s}^j})){h}({F}_{s}^j),
\end{split} 
\label{eq:fcsi}
\end{equation}
\\
where $f(\overline{F_c})=W_f\overline{F_c}$, $g(\overline{F_s})=W_g\overline{F_s}$, and $h(F_s)={W_h}{F_s}$. Further, $\overline{F}$ denotes a mean--variance channel-wise normalized version of $F$. The response is normalized by a factor ${C}({F})=\sum_{\forall{j}}\exp({f}(\overline{{F}_{c}^i})^T{g}(\overline{{F}_{s}^j}))$. Here, $i$ is the index of an output position and $j$ is the index that enumerates all possible positions. In the above formulation, $W_f$, $W_g$, and $W_h$ are the learned weight matrices, which are implemented as 1 $\times$ 1 convolutions as in \cite{zhang2018self}.

Our SANet has a network structure similar to the existing non-local block structure \cite{wang2017non}, but the number of input data differ (the input of the SANet consists of $F_c$ and $F_s$ ). The SANet module can appropriately embed a local style pattern in each position of the content feature maps by mapping a relationship (such as affinity) between the content and style feature maps through learning.
%-------------------------------------------------------------------------

\subsection{Full System}
As shown in Fig.~\ref{fig:overview}, we use the encoder (a pre-trained VGG-19 \cite{simonyan2014very}) to compute the loss function for training the SANet and decoder:
\begin{equation}
\begin{split} 
\mathcal{L}&=\lambda_{c}\mathcal{L}_{c}+\lambda_{s}\mathcal{L}_{s}+\mathcal{L}_{identity},
\end{split}
\label{eq:loss_all}
\end{equation}
where the composers of content, style, and identity loss are $\mathcal{L}_{c}$, $\mathcal{L}_{s}$, and $\mathcal{L}_{identity}$, respectively, and $\lambda_{c}$ and $\lambda_{s}$ are the weights of different losses. 

Similar to \cite{huang2017arbitrary}, the content loss is the Euclidean distance between the mean--variance channel-wise normalized target features, $\overline{{F}_{c}^{\texttt{r\_4\_1}}}$ and $\overline{{F}_{c}^{\texttt{r\_5\_1}}}$ and the mean--variance channel-wise normalized features of the output image VGG features, $\overline{{E}({I}_{cs})^{\texttt{r\_4\_1}}}$ and  $\overline{{E}({I}_{cs})^{\texttt{r\_5\_1}}}$, as follows:

\begin{equation}
\begin{split} 
\mathcal{L}_{c}&=||\overline{{E}({I}_{cs})^{\texttt{r\_4\_1}}}-\overline{{F}_{c}^{\texttt{r\_4\_1}}}||_2+||\overline{{E}({I}_{cs})^{\texttt{r\_5\_1}}}-\overline{{F}_{c}^{\texttt{r\_5\_1}}}||_{2}.
\end{split}
\label{eq:loss_content}
\end{equation}

The style loss is defined as follows:
\begin{equation}
\begin{split} 
\mathcal{L}_{s}&=\sum_{i=1}^L||{\mu}({\phi}_{i}({I}_{cs}))-{\mu}({\phi}_{i}({I}_{s}))||_{2}
\\&+||{\sigma}({\phi}_{i}({I}_{cs}))-{\sigma}({\phi}_{i}({I}_{s}))||_{2},
\end{split}
\label{eq:loss_style}
\end{equation}
where each ${\phi}$ denotes a feature map of the layer in the encoder used to compute the style loss. We use \texttt{Relu\_1\_1}, \texttt{Relu\_2\_1}, \texttt{Relu\_3\_1}, \texttt{Relu\_4\_1}, and \texttt{Relu\_5\_1} layers with equal weights. We have applied both the Gram matrix loss \cite{gatys2016image} and the AdaIN style loss \cite{huang2017arbitrary}, but the results show that the AdaIN style loss is more satisfactory.

When $W_f$, $W_g$, and $W_h$ are fixed as the identity matrices, each position in the content feature maps can be transformed into the semantically nearest feature in the style feature maps. In this case, the system cannot parse sufficient style features. In the SANet, although $W_f$, $W_g$, and $W_h$ are learnable matrices, our style transfer model can be trained by considering only the global statistics of the style loss $\mathcal{L}_{s}$.

To consider both the global statistics and the semantically local mapping between the content features and the style features, we define a new identity loss function as
\begin{equation}
\begin{split} 
\mathcal{L}_{identity}&=\lambda_{identity1}(||({I}_{cc}-{I}_{c})||_{2}+||({I}_{ss}-{I}_{s})||_{2})
\\&+\lambda_{identity2}\sum_{i=1}^L(||{\phi_i}({I}_{cc})-{\phi_i}({I}_{c})||_{2}
\\&+||{\phi_i}({I}_{ss})-{\phi_i}({I}_{s})||_{2}),
\end{split}
\label{eq:loss_identity}
\end{equation}
where $I_{cc}$(or $I_{ss}$) denotes the output image synthesized from two same content (or style) images, each ${\phi_i}$ denotes a layer in the encoder, and $\lambda_{identity1}$ and $\lambda_{identity2}$ are identity loss weights. The weighting parameters are simply set as $\lambda_{c}=1$, $\lambda_{s}=3$, $\lambda_{identity1}=1$, and $\lambda_{identity2}=50$ in our experiments.

The content and style losses control the trade-off between the structure of the content image and the style patterns. Unlike the other two losses, the identity loss is calculated from the same input images with no gap in style characteristics. Therefore, the identity loss concentrates on keeping the structure of the content image rather than changing the style statistics. As a result, the identity loss makes it possible to maintain the structure of the content image and style characteristics of the reference image simultaneously.

\begin{figure}
\begin{center}
\includegraphics[width=1\linewidth, height=0.5\linewidth]{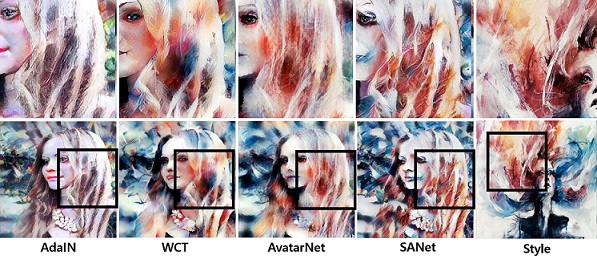}
\end{center}
  \caption{Result details. Regions marked by bounding boxes in the bottom row are enlarged in the top row for better visualization.}
\label{fig:closeup}
\end{figure}

\begin{figure}
\begin{center}
\includegraphics[width=1\linewidth, height=0.6\linewidth]{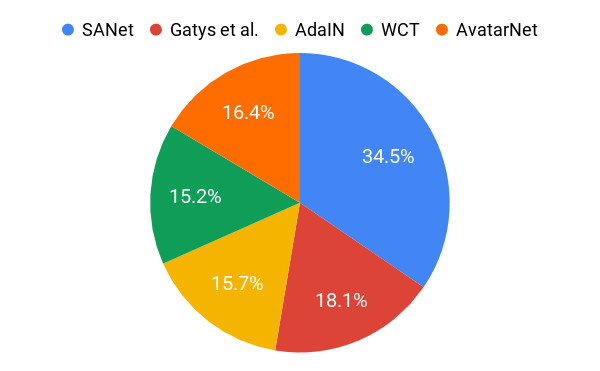}
\end{center}
  \caption{User preference result of five style transfer algorithms.}
\label{fig:userstudy}
\end{figure}

\begin{table}[]
\begin {center}
\begin{tabular}{c|cc}
Method & Time (256 px) & Time (512 px)\\ \hline
 Gatys et al. \cite{gatys2016image} & 15.863 & 50.804\\
 WCT \cite{li2017universal} & 0.689 & 0.997 \\
 Avatar-Net \cite{sheng2018avatar} & 0.248 & 0.356  \\
 AdaIN \cite{huang2017arbitrary} & 0.011 & 0.039  \\ \hline
 ours (\texttt{Relu\_4\_1}) & 0.012 & 0.042  \\
 ours (multi-level)& 0.017 & 0.055 
\end{tabular}
\end {center}
\caption{Execution time comparison (in seconds).}
\label{table:excution_time}
\end{table}

%-------------------------------------------------------------------------

\section{Experimental Results}

Figure~\ref{fig:overview} shows an overview of our style transfer network based on the proposed SANets. The demo site will be release at https://dypark86.github.io/SANET/.
%-------------------------------------------------------------------------
\subsection{Experimental Settings}
We trained the network using MS-COCO \cite{lin2014microsoft} for the content images and WikiArt \cite{phillips2011wiki} for the style images. Both datasets contain roughly 80,000 training images. We used the Adam optimizer \cite{kingma2014adam} with a learning rate of 0.0001 and a batch size of five content--style image pairs. During training, we first rescaled the smaller dimension of both images to 512 while preserving the aspect ratio, then randomly cropped a region of size 256$\times$256 pixels. In the testing phase, our network can handle any input size because it is fully convolutional.
%-------------------------------------------------------------------------
\subsection{Comparison with Prior Work}
To evaluate the our method, we compared it with three types of arbitrary style transform methods: the iterative optimization method proposed by Gatys et al. \cite{gatys2016image}, two feature transformation-based methods (WCT \cite{li2017universal} and AdaIN \cite{huang2017arbitrary}), and the patch-based method Avatar-Net \cite{sheng2018avatar}.

\begin{figure*}
\begin{center}
\includegraphics[width=1\linewidth, height=1.23\linewidth]{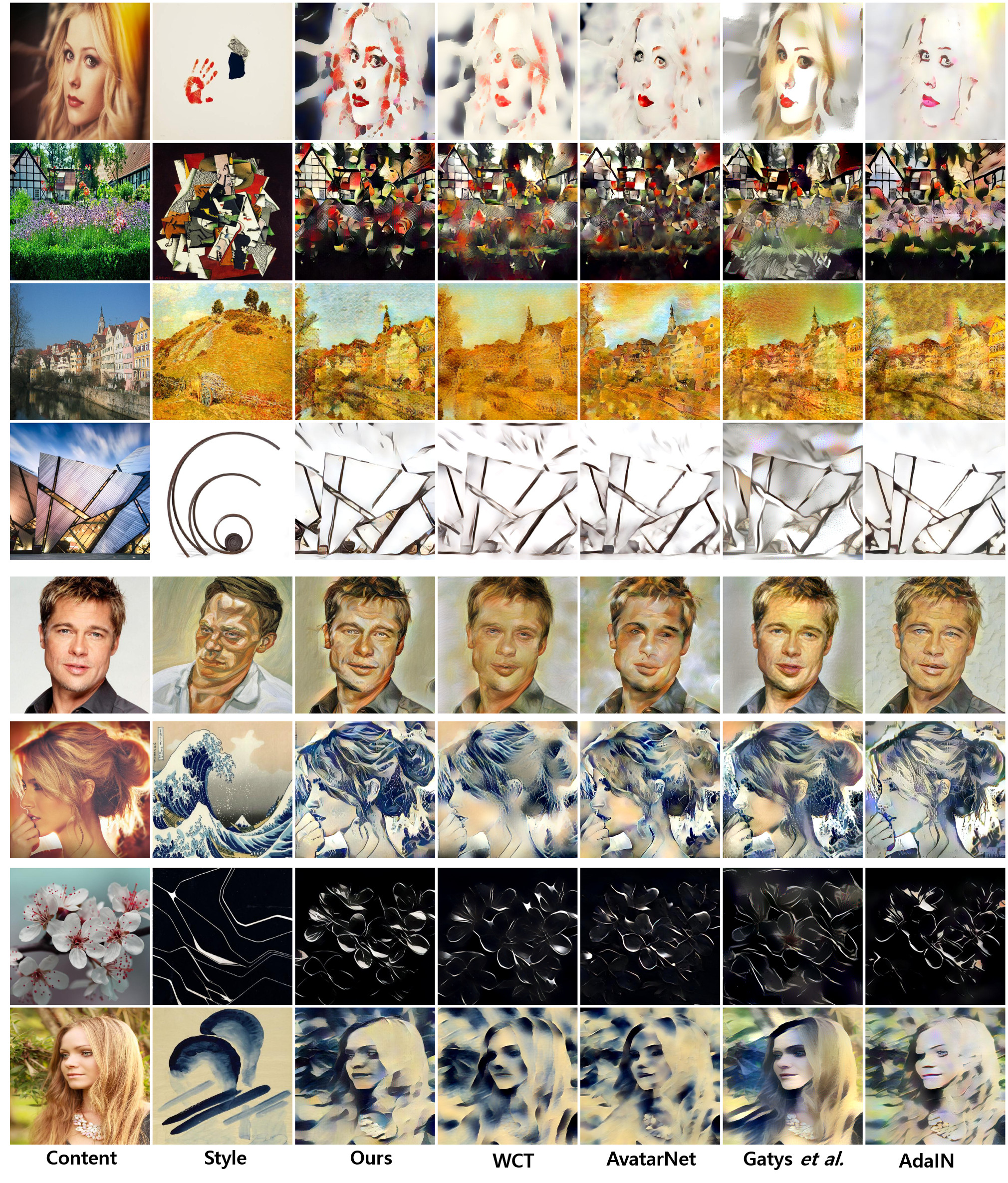}
\end{center}
  \caption{Example results for comparisons.}
\label{fig:comparisons}
\end{figure*}

{\bf Qualitative examples.} In Fig.~\ref{fig:comparisons}, we show examples of style transfer results synthesized by the state-of-the-art methods. Additional results are provided in the supplementary materials. Note that none of the test style images were observed during the training of our model. 

The optimization-based method \cite{gatys2016image} allows arbitrary style transfer, but is likely to encounter a bad local minimum (e.g., rows 2 and 4 in Fig.~\ref{fig:comparisons}). AdaIN \cite{huang2017arbitrary} simply adjusts the mean and variance of the content features to synthesize the stylized image. However, its results are less appealing and often retain some of the color distribution of the content due to the trade-off between content and style (e.g., rows 1, 2, and 8 in Fig.~\ref{fig:comparisons}). In addition, both AdaIN \cite{huang2017arbitrary} and WCT \cite{li2017universal} sometimes yield distorted local style patterns because of the holistic adjustment of the content features to match the second-order statistics of the style features, as shown in Fig.~\ref{fig:comparisons}. Although Avatar-Net \cite{sheng2018avatar} decorates the image with the style patterns according to the semantic spatial distribution of the content image and applies a multi-scale style transfer, it frequently cannot represent the local and global style patterns at the same time due to its dependency on the patch size. Moreover, it cannot keep the content structure in most cases (column 4 in Fig.~\ref{fig:comparisons}). In contrast, our method can parse diverse style patterns such as global color distribution, texture, and local style patterns while maintaining the structure of the content in most examples, as shown in Fig.~\ref{fig:comparisons}. 

Unlike other algorithms, our learnable SANets can flexibly parse a sufficient level of style features without maximally aligning the content and style features, regardless a large domain gap (rows 1 and 6 in Fig.~\ref{fig:comparisons}).   The proposed SANet semantically distinguishes the content structure and transfers similar style patterns onto the regions with the same semantic meaning. Our method transfers different styles for each type of semantic content. In Fig.~\ref{fig:comparisons} (row 3), the sky and buildings in our stylized image are stylized using different style patterns, whereas the results of other methods have ambiguous style boundaries between the sky and buildings. 

We also provide details of the results in Fig.~\ref{fig:closeup}. Our results exhibit multi-scale style patterns (e.g., color distribution, bush strokes, and the white and red patterns of rough textures in the style image). Avatar-Net and WCT distort the brush strokes, output blurry hair texture, and do not preserve the appearance of the face. AdaIN cannot even preserve the color distribution.

{\bf User study.} We used 14 content images and 70 style images to synthesize 980 images in total. We randomly selected 30 content and style combinations for each subject and showed them the stylized images obtained by the five comparison methods side-by-side in a random order. We then asked the subject to indicate his/her favorite result for each style. We collect 2,400 votes from 80 users and show the percentage of votes for each method in Fig.~\ref{fig:userstudy}. The result shows that the stylized results obtained by our method are preferred more often than those of other methods.

{\bf Efficiency.} Table~\ref{table:excution_time} shows the run time performance of the proposed method and other methods at two image scales: 256 and 512 pixels. We measured the runtime performance, including the time for style encoding. The optimization-based method \cite{gatys2016image} is impractically computationally expensive because of its iterative optimization process. In contrast, our multi-scale models (\texttt{Relu\_4\_1} and \texttt{Relu\_5\_1}) algorithms run at 59 fps and 18 fps for 256- and 512-pixel images respectively, and the single-scale (only \texttt{Relu\_4\_1}) algorithms runs at 83 fps and 24 fps for 256- and 512-pixel images respectively. Therefore, our method could feasibly process style transfer in real time. Our model is 7--20 times faster than the matrix computation-based methods (WCT \cite{li2017universal} and Avatar-Net \cite{sheng2018avatar}). 

\begin{figure}
\begin{center}
\includegraphics[width=1\linewidth, height=0.55\linewidth]{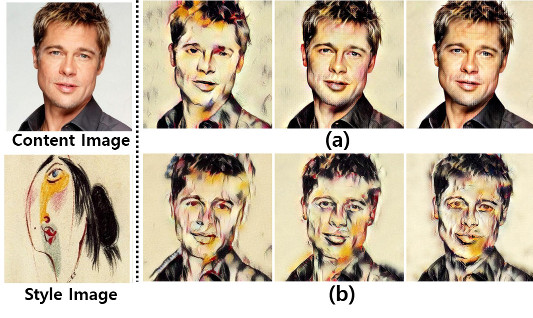}
\end{center}
  \caption{Content-style loss vs. identity loss. (a) Results obtained by fixing $\lambda_{identity1}$, $\lambda_{identity2}$, and $\lambda_{s}$ at 0, 0, and 5, respectively, and increasing $\lambda_{c}$ from 1 to 50. (b) Results obtained by fixing $\lambda_{c}$ and $\lambda_{s}$ at 0 and 5, respectively, and increasing $\lambda_{identity1}$ and $\lambda_{identity2}$ from 1 to 100 and from 50 to 5,000, respectively.}
\label{fig:identity_loss}
\end{figure}

\begin{figure}
\begin{center}
\includegraphics[width=1\linewidth, height=0.25\linewidth]{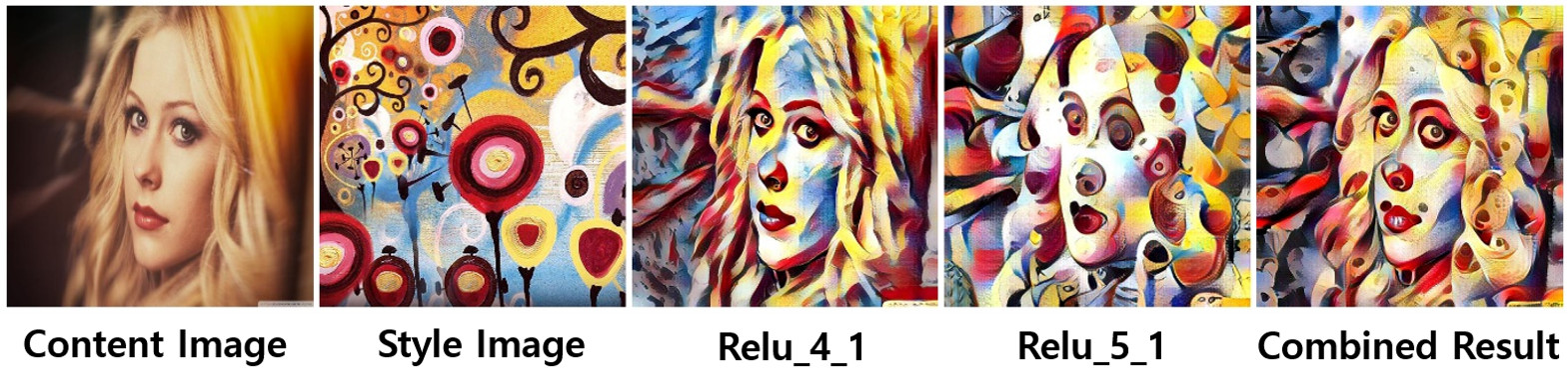}
\end{center}
  \caption{Multi-level feature embedding. By embedding features at multiple levels, we can enrich the local and global patterns for the stylized images.}
\label{fig:multi_level}
\end{figure}

%-------------------------------------------------------------------------
\subsection{Ablation Studies}

{\bf Loss analysis.} In this section, we show the influence of content-style loss and identity loss. Figure~\ref{fig:identity_loss} (a) shows the results obtained by fixing $\lambda_{identity1}$, $\lambda_{identity2}$, and $\lambda_{s}$ at 0, 0, and 5, respectively, while increasing $\lambda_{c}$ from 1 to 50. Figure~\ref{fig:identity_loss} (b) shows the results obtained by fixing $\lambda_{c}$ and $\lambda_{s}$ at 0 and 5, respectively, and increasing $\lambda_{identity1}$ and $\lambda_{identity2}$ from 1 to 100 and from 50 to 5,000, respectively. Without the identity loss, if we increase the weight of the content loss, the content structure is preserved, but the characteristics of the style patterns disappear, because of the trade-off between the content loss and the style loss. In contrast, increasing the weights of identity loss without content loss preserves the content structure as much as possible while maintaining style patterns. However, distortion of the content structure cannot be avoided. We hence applied a combination of content-style loss and identity loss to maintain the content structure while enriching style patterns.

\begin{figure*}
\begin{center}
\includegraphics[width=1\linewidth, height=0.2\linewidth]{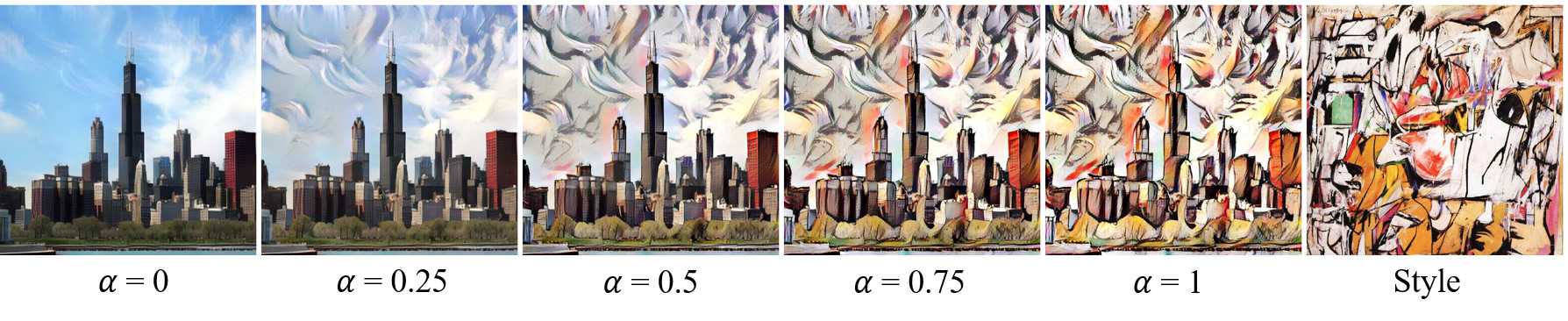}
\end{center}
  \caption{Content--style trade-off during runtime. Our algorithm allows this trade-off to be adjusted at runtime by interpolating between feature maps ${F}_{ccc}^m$ and ${F}_{csc}^m$.}
\label{fig:tradeoff}
\end{figure*}

\begin{figure}
\begin{center}
\includegraphics[width=1\linewidth, height=0.8\linewidth]{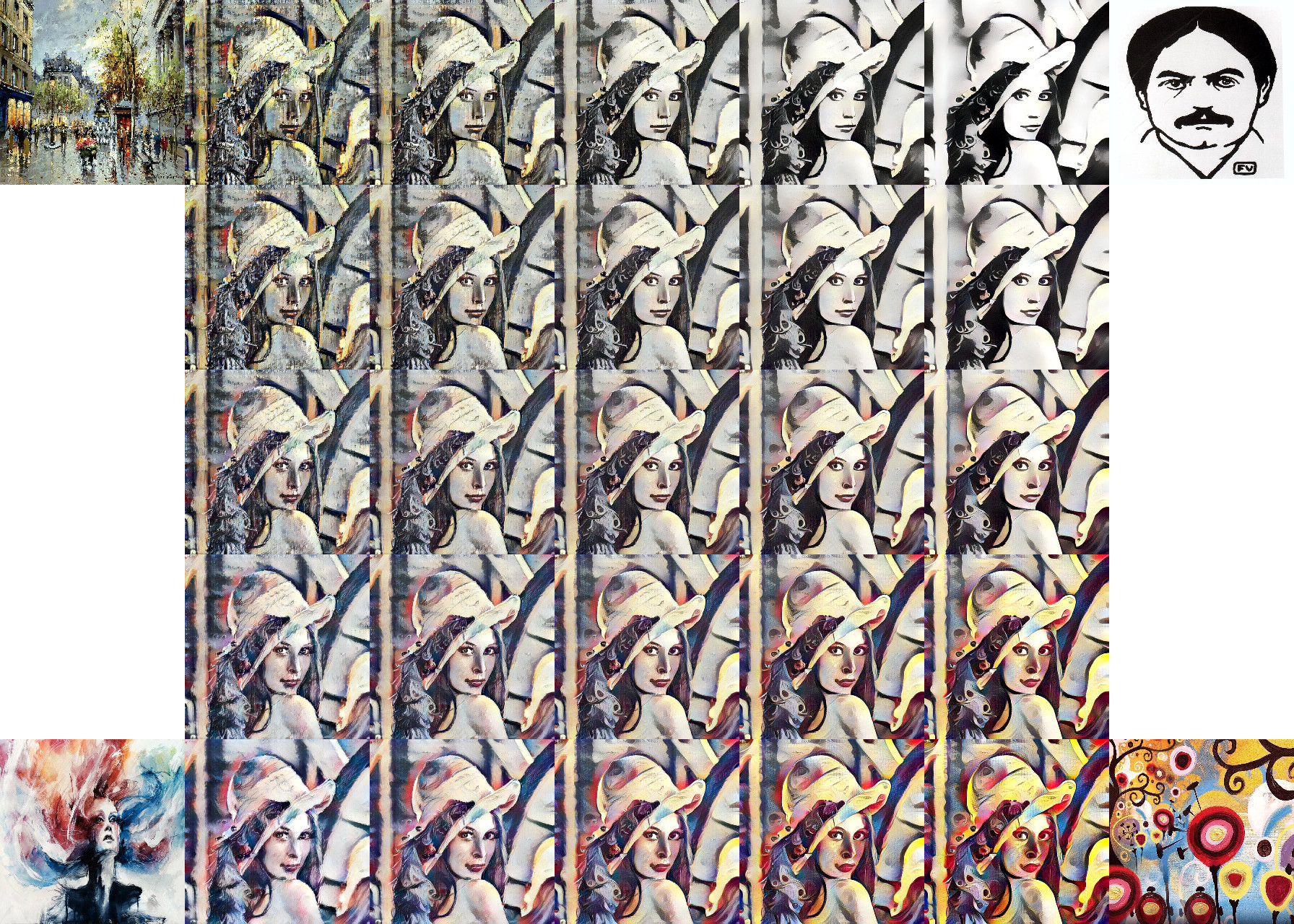}
\end{center}
  \caption{Style interpolation with four different styles.}
\label{fig:interpolation}
\end{figure}

\begin{figure}
\begin{center}
\includegraphics[width=1\linewidth, height=0.4\linewidth]{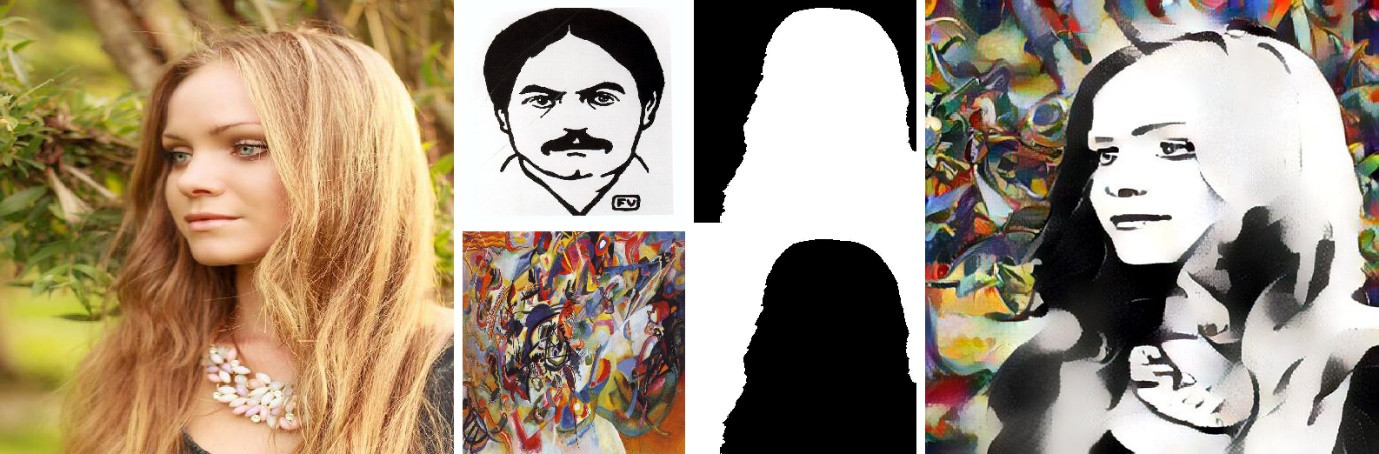}
\end{center}
  \caption{Example of spatial control. Left: content image. Middle: style images and masks. Right: stylized image from two different style images.}
\label{fig:spatial_control}
\end{figure}

{\bf Multi-level feature embedding.} Figure~\ref{fig:multi_level} shows two stylized outputs obtained from \texttt{Relu\_4\_1} and \texttt{Relu\_5\_1}, respectively. When only \texttt{Relu\_4\_1} is used for style transfer, the global statistics of the style features and the content structure are maintained well. However, the local style patterns do not appear well. In contrast, \texttt{Relu\_5\_1} helps add the local style patterns such as circle patterns because the receptive field is wider. However, the content structures are distorted and textures such as brush strokes disappear. In our work, to enrich the style patterns, we integrated two SANets by taking VGG feature maps encoded from the different (\texttt{Relu\_4\_1} and \texttt{Relu\_5\_1}) layers as inputs and combining both output feature maps

%-------------------------------------------------------------------------

\subsection{Runtime Controls}
In this section, we present the flexibility of our method through several applications.

{\bf Content--style trade-off.} The degree of stylization can be controlled during training by adjusting the style weight $\lambda_s$ in Eq.~\ref{eq:loss_all} or during test time by interpolating between feature maps that are fed to the decoder. For runtime control, we adjust the stylized features ${F}_{csc}^m\xleftarrow{}\alpha{F}_{csc}^m+(1-\alpha){F}_{ccc}^m$ and $\forall{\alpha}\in[0,1]$. Map ${F}_{ccc}^m$ is obtained by taking two content images as input for our model. The network tries to reconstruct the content image when $\alpha = 0$, and to synthesize the most stylized image when $\alpha = 1$ (as shown in Fig.~\ref{fig:tradeoff}). 

{\bf Style interpolation.} To interpolate between several style images, a convex combination of feature maps ${F}_{csc}^m$ from different styles can be fed into the decoder (as shown in Fig.~\ref{fig:interpolation}).

{\bf Spatial control.} Figure~\ref{fig:spatial_control} shows an example of spatially controlling  the stylization. A set of masks $M$ (Fig.~\ref{fig:spatial_control} column 3)  is additionally required as input to map the spatial correspondence between content regions and styles. We can assign the different styles in each spatial region by replacing ${F}_{csc}^m$ with ${M}\bigodot{F}_{csc}^m$, where $\bigodot$ is a simple mask-out operation.

\section{Conclusions}
In this work, we proposed a new arbitrary style transform algorithm that consists of style-attentional networks and decoders. Our algorithm is effective and efficient. Unlike the patch-based style decorator in \cite{sheng2018avatar}, our proposed SANet can flexibly decorate the style features through learning using a conventional style reconstruction loss and identity loss. Furthermore, the proposed identity loss helps the SANet maintain the content structure, enriching the local and global style patterns. Experimental results demonstrate that the proposed method synthesizes images that are preferred over other state-of-the-art arbitrary style transfer algorithms.

{\bf Acknowledgments.} This research is supported by the Ministry of Culture, Sports, and Tourism (MCST) and Korea Creative Content Agency (KOCCA) in the Culture Technology (CT) Research and Development Program 2019

%-------------------------------------------------------------------------
{\small
\bibliographystyle{ieee}
%\bibliography{egbib}

\begin{thebibliography}{10}\itemsep=-1pt

\bibitem{chen2017stylebank}
D.~Chen, L.~Yuan, J.~Liao, N.~Yu, and G.~Hua.
\newblock StyleBank: An explicit representation for neural image style
  transfer.
\newblock In {\em Proc. CVPR}, volume~1, page~4, 2017.

\bibitem{chen2016fast}
T.~Q. Chen and M.~Schmidt.
\newblock Fast patch-based style transfer of arbitrary style.
\newblock {\em arXiv preprint arXiv:1612.04337}, 2016.

\bibitem{dumoulin2017learned}
V.~Dumoulin, J.~Shlens, and M.~Kudlur.
\newblock A learned representation for artistic style.
\newblock In {\em Proc. ICLR}, 2017.

\bibitem{gatys2015texture}
L.~Gatys, A.~S. Ecker, and M.~Bethge.
\newblock Texture synthesis using convolutional neural networks.
\newblock In {\em Advances in Neural Information Processing Systems}, pages
  262--270, 2015.

\bibitem{gatys2016image}
L.~A. Gatys, A.~S. Ecker, and M.~Bethge.
\newblock Image style transfer using convolutional neural networks.
\newblock In {\em Proc. CVPR}, pages 2414--2423, 2016.

\bibitem{gatys2017controlling}
L.~A. Gatys, A.~S. Ecker, M.~Bethge, A.~Hertzmann, and E.~Shechtman.
\newblock Controlling perceptual factors in neural style transfer.
\newblock In {\em Proc. CVPR}, 2017.

\bibitem{huang2017arbitrary}
X.~Huang and S.~J. Belongie.
\newblock Arbitrary style transfer in real-time with adaptive instance
  normalization.
\newblock In {\em Proc. ICCV}, pages 1510--1519, 2017.

\bibitem{johnson2016perceptual}
J.~Johnson, A.~Alahi, and L.~Fei-Fei.
\newblock Perceptual losses for real-time style transfer and super-resolution.
\newblock In {\em Proc. ECCV}, pages 694--711.
  Springer, 2016.

\bibitem{kingma2014adam}
D.~P. Kingma and J.~Ba.
\newblock Adam: A method for stochastic optimization.
\newblock {\em arXiv preprint arXiv:1412.6980}, 2014.

\bibitem{li2016combining}
C.~Li and M.~Wand.
\newblock Combining Markov random fields and convolutional neural networks for
  image synthesis.
\newblock In {\em Proc. CVPR}, pages 2479--2486, 2016.

\bibitem{li2016precomputed}
C.~Li and M.~Wand.
\newblock Precomputed real-time texture synthesis with Markovian generative
  adversarial networks.
\newblock In {\em Proc. ECCV}, pages 702--716.
  Springer, 2016.

\bibitem{li2017diversified}
Y.~Li, C.~Fang, J.~Yang, Z.~Wang, X.~Lu, and M.-H. Yang.
\newblock Diversified texture synthesis with feed-forward networks.
\newblock In {\em Proc. CVPR}, 2017.

\bibitem{li2017universal}
Y.~Li, C.~Fang, J.~Yang, Z.~Wang, X.~Lu, and M.-H. Yang.
\newblock Universal style transfer via feature transforms.
\newblock In {\em Advances in Neural Information Processing Systems}, pages
  386--396, 2017.

\bibitem{li2017demystifying}
Y.~Li, N.~Wang, J.~Liu, and X.~Hou.
\newblock Demystifying neural style transfer.
\newblock {\em arXiv preprint arXiv:1701.01036}, 2017.

\bibitem{lin2014microsoft}
T.-Y. Lin, M.~Maire, S.~Belongie, J.~Hays, P.~Perona, D.~Ramanan,
  P.~Doll{\'a}r, and C.~L. Zitnick.
\newblock Microsoft COCO: Common objects in context.
\newblock In {\em Proc. ECCV}, pages 740--755.
  Springer, 2014.

\bibitem{paszkepytorch}
A.~Paszke, S.~Chintala, R.~Collobert, K.~Kavukcuoglu, C.~Farabet, S.~Bengio,
  I.~Melvin, J.~Weston, and J.~Mariethoz.
\newblock PyTorch: Tensors and dynamic neural networks in Python with strong
  GPU acceleration, Available: https://github.com/pytorch/pytorch, May 2017.

\bibitem{phillips2011wiki}
F.~Phillips and B.~Mackintosh.
\newblock Wiki Art Gallery, Inc.: A case for critical thinking.
\newblock {\em Issues in Accounting Education}, 26(3):593--608, 2011.

\bibitem{risser2017stable}
E.~Risser, P.~Wilmot, and C.~Barnes.
\newblock Stable and controllable neural texture synthesis and style transfer
  using histogram losses.
\newblock {\em arXiv preprint arXiv:1701.08893}, 2017.

\bibitem{shen2017meta}
F.~Shen, S.~Yan, and G.~Zeng.
\newblock Meta networks for neural style transfer.
\newblock {\em arXiv preprint arXiv:1709.04111}, 2017.

\bibitem{sheng2018avatar}
L.~Sheng, Z.~Lin, J.~Shao, and X.~Wang.
\newblock Avatar-Net: Multi-scale zero-shot style transfer by feature
  decoration.
\newblock In {\em Proc. CVPR}, pages 8242--8250, 2018.

\bibitem{simonyan2014very}
K.~Simonyan and A.~Zisserman.
\newblock Very deep convolutional networks for large-scale image recognition.
\newblock {\em arXiv preprint arXiv:1409.1556}, 2014.

\bibitem{ulyanov2016texture}
D.~Ulyanov, V.~Lebedev, A.~Vedaldi, and V.~S. Lempitsky.
\newblock Texture networks: Feed-forward synthesis of textures and stylized
  images.
\newblock In {\em Proc. ICML}, pages 1349--1357, 2016.

\bibitem{ulyanovinstance}
D.~Ulyanov, A.~Vedaldi, and V.~Lempitsky.
\newblock Instance normalization: The missing ingredient for fast stylization.
\newblock {\em arXiv preprint arXiv:1607.08022}, (2016).

\bibitem{ulyanov2017improved}
D.~Ulyanov, A.~Vedaldi, and V.~S. Lempitsky.
\newblock Improved texture networks: Maximizing quality and diversity in
  feed-forward stylization and texture synthesis.
\newblock In {\em Proc. CVPR}, volume~1, page~3, 2017.

\bibitem{vaswani2017attention}
A.~Vaswani, N.~Shazeer, N.~Parmar, J.~Uszkoreit, L.~Jones, A.~N. Gomez,
  {\L}.~Kaiser, and I.~Polosukhin.
\newblock Attention is all you need.
\newblock In {\em Advances in Neural Information Processing Systems}, pages
  5998--6008, 2017.

\bibitem{wang2017zm}
H.~Wang, X.~Liang, H.~Zhang, D.-Y. Yeung, and E.~P. Xing.
\newblock ZM-Net: Real-time zero-shot image manipulation network.
\newblock {\em arXiv preprint arXiv:1703.07255}, 2017.

\bibitem{wang2017non}
X.~Wang, R.~Girshick, A.~Gupta, and K.~He.
\newblock Non-local neural networks.
\newblock {\em arXiv preprint arXiv:1711.07971}, 2017.

\bibitem{wang2017multimodal}
X.~Wang, G.~Oxholm, D.~Zhang, and Y.-F. Wang.
\newblock Multimodal transfer: A hierarchical deep convolutional neural network
  for fast artistic style transfer.
\newblock In {\em Proc. CVPR}, volume~2, page~7, 2017.

\bibitem{zhang2017multi}
H.~Zhang and K.~Dana.
\newblock Multi-style generative network for real-time transfer.
\newblock {\em arXiv preprint arXiv:1703.06953}, 2017.

\bibitem{zhang2018self}
H.~Zhang, I.~Goodfellow, D.~Metaxas, and A.~Odena.
\newblock Self-attention generative adversarial networks.
\newblock {\em arXiv preprint arXiv:1805.08318}, 2018.

\end{thebibliography}

}

\end{document}